\documentclass[utf8]{FrontiersinHarvard} 

\usepackage{url,hyperref,lineno,microtype,subcaption}
\usepackage[onehalfspacing]{setspace}

\usepackage{nameref,hyperref}
\usepackage[table]{xcolor}
\hypersetup{
    colorlinks = true, 
    citecolor = {blue},
    linkcolor = {teal},
           }

\def\keyFont{\fontsize{8}{11}\helveticabold }

\def\firstAuthorLast{Gros {et~al.}} 
\def\Authors{Claudius Gros\,$^{1,*}$}

\def\Address{$^{1}$Institute for Theoretical Physics, Goethe
University, Frankfurt a.M., Germany 
}

\def\corrAuthor{Claudius Gros} 
\def\corrEmail{gros07@itp.uni-frankfurt.edu} 

\begin{document}
\onecolumn
\firstpage{1}

\title[Takeaways from generative AI]{
From generative AI to the brain: 
five takeaways}

\author[\firstAuthorLast ]{\Authors}
\address{} 
\correspondance{} 
\extraAuth{} 

\maketitle

\begin{abstract}
The big strides seen in generative AI are not 
based on somewhat obscure algorithms, but due 
to clearly defined generative principles. The
resulting concrete implementations have proven 
themselves in large numbers of applications. 
We suggest that it is imperative to thoroughly 
investigate which of these generative principles 
may be operative also in the brain, and hence 
relevant for cognitive neuroscience. In addition, 
ML research led to a range of interesting 
characterizations of neural information 
processing systems. We discuss five examples, 
the shortcomings of world modelling, the 
generation of thought processes, attention, 
neural scaling laws, and quantization,
that illustrate how much neuroscience could
potentially learn from ML research. 
\section{}

\tiny
\keyFont{ \section{Keywords:} generative AI, 
cognitive neuroscience, attention, 
predictive coding, chain of thought, quantization} 
\end{abstract}

\section{Introduction}
A multitude of factors contributes to the current
rise of generative artificial intelligence
(generative AI). Here we focus on two aspects.
\begin{itemize}
\item Algorithmic developments can be formulated 
in many cases in terms of generic generative principles.
These generative principles have proven themselves,
giving rise to high-performing machine learning 
architectures. It is an important question whether 
corresponding principles may operate in the 
brain.
\item In addition to algorithms, insights regarding 
general working principles and properties of 
neural-based information processing systems have 
been attained. Do these apply also
to the human brain?
\end{itemize}
Machine learning (ML) offers a range of
conjectures for the workings of our brain, 
some of which extend or parallel traditional
neuroscience frameworks, while others are   
new. Cognitive neuroscience should accept 
the challenge and evaluate these conjectures 
systematically in the context of wet information 
processing.

A comprehensive overview of potentially relevant 
cross-relations between ML and the neurosciences
is beyond the scope of this perspective. We will
focus instead on five key aspects elucidating
the importance of paying attention to the
concepts that are being developed for generative 
artificial intelligence. A flurry of new ideas 
awaits the scrutiny of
cognitive neuroscience.

\section{World modelling is not enough}

The two learning principles, `predictive coding' 
(neuroscience) and `autoregressive language modeling' 
are both dedicated to the task of building
world models, with the former also having 
active components \citep{brodski2017information}, 
operating in addition
on distinct scales and modalities
\citep{caucheteux2023evidence}.

For large language models (LLMs), autoregressive 
language modeling takes the form of next-word 
predictions.
However, ML tells us that world-model building
alone is insufficient.
 
\medskip
\centerline{\setlength\arrayrulewidth{2pt}
\def\arraystretch{1.2}
\begin{tabular}{|l|}
\parbox{0.8\textwidth}{
The base- or foundation model, viz
the result of word prediction training, 
does contain the knowledge of the world, 
as present in the training data. 
But all it can do is to complete a given 
input word by word. At this stage, 
key concepts of relevance for the interaction 
with users, such as `question' and `answer', 
are not yet explicitly encoded.}
\end{tabular}}

\medskip
In the early 2020s, a significant step 
forward was the realization that the
otherwise essentially useless base model 
can be turned into a cognitive powerhouse 
via a secondary process, denoted `fine tuning' 
or `human supervised fine-tuning' (HSFT). 
\begin{itemize}
\item A core fine tuning objective is to teach the 
system to generate meaningful responses for a given
prompt, and not just engage in text completion.
\item Next comes fine tuning of style, political 
correctness, etc.
\item Models may be fine tuned further for specific
downstream tasks, specializing the otherwise universal
LLM to excel, e.g., in accounting.
\end{itemize}
It seems likely that equivalent processes 
would occur in our brains. In ML, the two 
processes are normally separated, viz 
performed subsequently. In the brain, world model 
training and fine-tuning via reinforcements are 
conceivably active at the same time.

{\bf Takeaway:}\ ML offers a concrete 
construction plan for a basic cognitive system: 
Universal unsupervised world modelling 
followed by supervised fine tuning. To 
which extent does the brain follow this recipe?

\section{Generative principles for human thinking}

The autonomous generation of thoughts 
is considered to be the basis of human 
intelligence. It is hence remarkable
that commercial chatbots started to engage in 
rudimentary `thinking' by the mid-2020s. 
The algorithm used is denoted 
'Chain-of-Thought' (CoT) \citep{zhang2025igniting}, 
originally a prompting technique 
\citep{wei2022chain}.
It is unclear to which extent human thought 
processes may be understood within the
CoT framework, if at all. The same
holds for its generalizations,
viz `Chain-of-X' (CoX) \citep{xia2024beyond},
such as Chain-of-Feedback, 
Chain-of-Instructions, or Chain-of-Histories.
In any case, of interest are the underlying 
generative principles.
\begin{itemize}
\item CoT is one of many possible fine-tuning 
processes, characterized by a specific
objective function.
\item The system auto-prompts, appending its own 
thoughts to the user prompt.
\item The response is then generated using the combined
prompt: (user input)+(chain of thoughts). 
\end{itemize}
Why are responses substantially better when 
the LLM thinks for a while? A possible 
explanation is based on the
information bottleneck (IB) framework
\citep{tishby2015deep}. We recall
that the token sequence is
$$
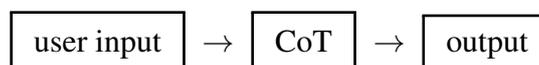
\fbox{\,user\ input\,} \ \to\ \fbox{\,CoT\,}
\ \to\ \fbox{\,output\,}
$$
The middle part, the thought processes, can
be interpreted to act as an information 
bottleneck for the cognitive processing 
between input and output 
\citep{lei2025revisiting}. This
principle can be expressed as an
information-theoretical min-max optimization.
\begin{itemize}
\item The mutual information between the input 
      and CoT is --minimized--.\\
This means that the self-generated thoughts should
abstract from the specific formulation of the input, 
retaining only the overall content.
\item The mutual information between CoT and the
      output is --maximized--.\\
This because the latent space, namely the thoughts, 
should be maximally informative with regard to the 
final output.
\end{itemize}
The IB view is not just an abstract cookbook.
Instead, it has proven itself as a 
high-performing training algorithm 
\citep{lei2025revisiting}. 

As an alternative to the notion of an
information bottleneck, it has been 
proposed that CoT-type thinking 
may be seen as an effort to build a 
composite object, the final response 
\citep{zhu2025flowrl}. 
This interpretation allows to leverage 
state-of-the-art algorithms for diverse 
object generation, such as GFlowNet 
\citep{bengio2021flow}, which is used
widely in synthetic chemistry.

{\bf Takeaway:}
The new views of the functionality of 
thought processes arising within ML research
should motivate us to pose one of the most 
fundamental questions the neurosciences
could consider. Could this provide a possible 
first step towards an understanding of human 
thinking?

\section{No attention without self-consistency}

A main driver of generative AI is the
self-attention mechanism powering the
transformer architecture
\citep{vaswani2017attention,de2022attention}.
Regarding the brain, we do not touch here 
the phenomenology of human attention, or 
the sometimes controversially discussed question 
how attention should be defined operatively
in psychology and in the neurosciences
\citep{hommel2019no,wu2024we}. Given
this caveat, a few comments can be made:
\begin{itemize}
\item The details of how attention works
on the level of individual neurons are 
generally not well understood \citep{moore2017neural}.

\item Top-down attention involves the modulation
of sensory processing areas by signals generated 
in higher brain regions. It is generally assumed 
that these modulatory processes depend only
on the specific attention signals, viz without 
being coupled to the actual process 
generating the top-down signal in the first place.

\item Bottom-up attention is observed when early 
areas react to salient features in the sensory 
input stream. Pop-out features are then forwarded 
with higher intensity
\citep{connor2004visual}.
\end{itemize}
Bottom-up attention could be interpreted
as a variant of self-attention, albeit with
a reduced dynamic range. The latter because 
lateral saliency detection evolves
only slowly in the adult brain
\citep{hopfinger2017introduction}.

Operatively, there is a key difference between 
the current view of top-down and ML attention. 
In the neurosciences, attention processes 
operating in early brain regions are investigated
separately from the question of how the
modulating top-down signals are generated 
via cognitive control \citep{badre2024cognitive}, 
e.g., in the context of cholinergic signaling
\citep{parikh2020cholinergic}.
No such separation is present in machine 
learning, for a good reason. Components of
larger models develop their own neural
language when trained separately
\citep{luduena2013self}; models
need therefore to be trained in 
their entirety for the individual components 
to be able to talk to each other.

The context window of a transformer 
represents past states, which implies
that the self-attention mechanism
discussed above operates in the time
domain, involving hence working-memory
aspects \citep{hintzman1984minerva}.
This connection has been addressed in 
the context of modern Hopfield networks
\citep{ramsauer2020hopfield,ororbia2023neuro}.

{\bf Takeaway:}
A full understanding of attention needs
to include the self-consistency loop
between the generation of attention signals
and their subsequent processing.

\section{Neural scaling laws}

Neural scaling laws describe how performance 
and training times scale with model and/or
data size \citep{kaplan2020scaling,hoffmann2022training,michaud2023quantization}.

As an example consider the relative 
performance of two fully trained models
(A and B), which are identical in all aspects, 
apart from model sizes. In good approximation
the relative performance is then 
\citep{neumann2022scaling,neumann2024alphazero}
\begin{equation}
\frac{P_A}{P_A+P_B}\ \sim\ \frac{N_A}{N_A+N_B}\,.
\label{P_A_B}
\end{equation}
where $N_A$ ($N_B$) are the respective numbers 
of adaptable parameters, and $P_A$ ($P_B$) the 
corresponding performances. In addition, 
larger systems need 
longer to train. For the training compute $C$,
viz for the total amount of resources (chips, 
time, ...) needed to train a model, one finds
a quadratic scaling relation,
$$
    C\ \sim\  N^2\,.
$$
This leads to an interesting hypothesis 
regarding putative limitations to the 
phylogenetic growth of the brain.  Assume 
we have two humans, H1 and H2, the first 
with a standard brain size, the second with 
a brain twice as large. If we take 15 years 
as the time to train standard human brains, 
H2 would need $2^2=4$ times as long, namely 
60 years. After 60 years of growing up, H2 
would have higher cognitive capabilities than 
H1, as given by (\ref{P_A_B}). However, ML
finds that performance remains somewhat flat
during most training, increasing rapidly only 
at later stages. This implies that H2 would
underperform H1 for extended periods, say the 
first 40-50 years of adolescence. Evolutionary 
speaking, doubling brain size may hence not be 
a viable option. Of course other limiting
factors, like metabolic costs, may have determined 
the size of our brains.

{\bf Takeaway:}\ Given that information 
processing networks characterize not only
modern machine learning architectures,
but also the brain, the biological implications 
of neural scaling deserve to be investigated.

\section{Quantization}

Large models have large numbers of adaptable
parameters, which one needs not only to store,
but also to keep in working memory, ready for
subsequent use. Typical floating point datatypes
are 32bits (or 64bits for double precision). In order
to save memory, and to make operations faster,
data sizes have been reduced in recent years
\citep{wei2024advances,gong2025survey}.
Currently, INT4 (4 bits) devices are being rolled 
out. For INT4, one has just $2^4=16$ possible values. 
Synaptic weights are hence `quantized', taking 
only one out of 16 possible states. Specialized 
GPUs support the involved operations efficiently.
In analogy, synaptic strength is quantized
also in the brain \citep{petersen1998all,liu2017gradation},
with the exact number of expressed states 
being debated.

{\bf Takeaway:}\ The computational consequences
of synapse quantization are well understood for
artificial neural nets. This knowledge should
be readily transferable to their biological
counterparts.

\section{Conclusions}

We reviewed five selected concepts
contributing to the rapid progress
of generative AI. Interestingly, in
machine learning literature, their relevance
to biological information processing systems
is rarely discussed, if at all, with 
attention being in part an exception
\citep{lindsay2020attention}. The scope of 
this perspective is to raise awareness that
a treasure of generative principles may
be hidden in ML literature.

\section*{Conflict of Interest Statement}
The authors declare that the research was 
conducted in the absence of any commercial 
or financial relationships that could be 
construed as a potential conflict of interest.

\section*{Author Contributions}
Equal contributions.

\section*{Funding}
This research received no external funding.

\section*{Acknowledgments}

This article benefitted from discussions with
Christian Fiebach, Ricardo Kienitz, and
B\'ulcsu S\'andor.


\end{document}